\documentclass{article}
\usepackage{kr}

\usepackage{times}
\usepackage{soul}
\usepackage{url}
\usepackage[hidelinks]{hyperref}
\usepackage[utf8]{inputenc}
\usepackage[small]{caption}
\usepackage{graphicx}
\usepackage{amsmath}
\usepackage{amsthm}
\usepackage{booktabs}
\usepackage{algorithm}
\usepackage{algorithmic}
\usepackage{todonotes}
\usepackage{color, colortbl}

\title{An Experimental Study of Formula Embeddings \\for Automated Theorem Proving in First-Order Logic}

\author{
\small 
Ibrahim Abdelaziz$^\ddag$\footnote{Equal Contribution}\and
Veronika Thost$^{\diamond*}$\and
Maxwell Crouse$^\S$\and
Achille Fokoue$^\ddag$
\affiliations
{\normalsize \rm $^\ddag$IBM Research~~~~~} 
{\normalsize \rm $^\diamond$MIT-IBM Watson AI Lab~~~~~~} 
{\normalsize \rm $^\S$Northwestern University}
\emails
\normalsize \rm ibrahim.abdelaziz1@ibm.com,
veronika.thost@ibm.com,
mvcrouse@u.northwestern.edu,
achille@us.ibm.com
}

\begin{document}
\maketitle

\begin{abstract}
Automated theorem proving in first-order logic is an active research area which is successfully supported by machine learning.
While there have been various proposals for encoding logical formulas into numerical vectors -- from simple strings to more involved graph-based embeddings -- 
little is known about how these different encodings compare.
In this paper, we study and experimentally compare pattern-based embeddings that are applied in current systems with popular graph-based encodings, most of which have not been considered in the theorem proving context before.
Our experiments show that the advantages of simpler encoding schemes in terms of runtime are outdone by more complex graph-based embeddings, which yield more efficient search strategies and simpler proofs.
To support this, we present a detailed analysis across several dimensions of theorem prover performance beyond just proof completion rate, thus providing empirical evidence to help guide future research on neural-guided theorem proving towards the most promising directions.
\end{abstract}

\section{Introduction}
\label{sec:introduction}


First-order logic (FOL) theorem proving is important in many application domains \cite{Denney+-04-nasasw,Schumann-01-swe}.
State-of-the-art automated theorem provers~(ATPs) excel at finding complex proofs in restricted domains, 
but they have difficulty when reasoning in broader contexts; for example, with common sense knowledge and large mathematical libraries. Recently, the latter have become available in the form of logical theories (i.e., collections of axioms), and thus for reasoning \cite{grabowski2010mizar}. The challenge is now to extend traditional automated theorem provers to cope with the computational challenges inherent to reasoning at scale.

The theorem proving problem is as follows: given a set of axioms (formulas known to be true) and a conjecture formula, provide a proof for the conjecture that is derivable from the given axioms (if such a proof exists).
Classical algorithms for automated theorem proving usually rely on custom, manually designed 
heuristics based on analyses of formulas \cite{sekar2001term,hoder2011sine}. 
Several machine-learning based techniques have been shown recently to outperform or achieve competitive performance when compared to traditional heuristic-based methods~\cite{alama2014premise}, but they still depend on carefully selected manual features. 
Current research therefore focuses on the development of neural approaches \cite{bansal2019holist,Chvalovsky+-CADE19-enigmang,Crouse2019-trail}, 
which have more potential to lessen the need for the domain-specific feature engineering required by other techniques.
However, due to their highly structural and semantic nature, representing formulas for use with neural methods has been challenging.
The formula representations proposed in literature vary greatly: from rather simple approaches based on strings or sub-terms \cite{Alemi+-NIPS16-deepmath}, over more complex patterns \cite{JU-CICM17-enigma,Crouse2019-trail}, to encodings based on Tree LSTMs \cite{Loos+-LPAR2017-deep-nw-guided} and graph neural networks \cite{Crouse2019-premise-selection,paliwal2019graph}. 
%

Consider, the following example formula: 
$\forall A, B, C .\ r(A, B)\wedge\big(p(A) \vee \lnot q(B, f(A)) \vee q(C, f(A))\big)$, which is already in conjunctive normal form (CNF). 
The most simple embedding approaches consider the formula as a sequence of characters and use one-hot encodings followed by standard sequence encodings like LSTMs  \cite{Alemi+-NIPS16-deepmath}. This encoding does not capture much logical information, even syntactically. 
For example, that $f(A)$ occurs twice in different contexts represents two different  constraints on the interpretation of $A$, which should be reflected in the embedding of $A$. \cite{Alemi+-NIPS16-deepmath} therefore also propose a word-level encoding based on an iterative combination of the former embeddings.
Still, simple logical properties like the commutativity of $\vee$, meaning that the order of the literals $\neg q(B,f(A))$ and $q(C,f(A))$ is not relevant, would be extremely difficult to capture with a sequence-based approach.

\begin{figure}
\begin{centering}
\includegraphics[width=\columnwidth]{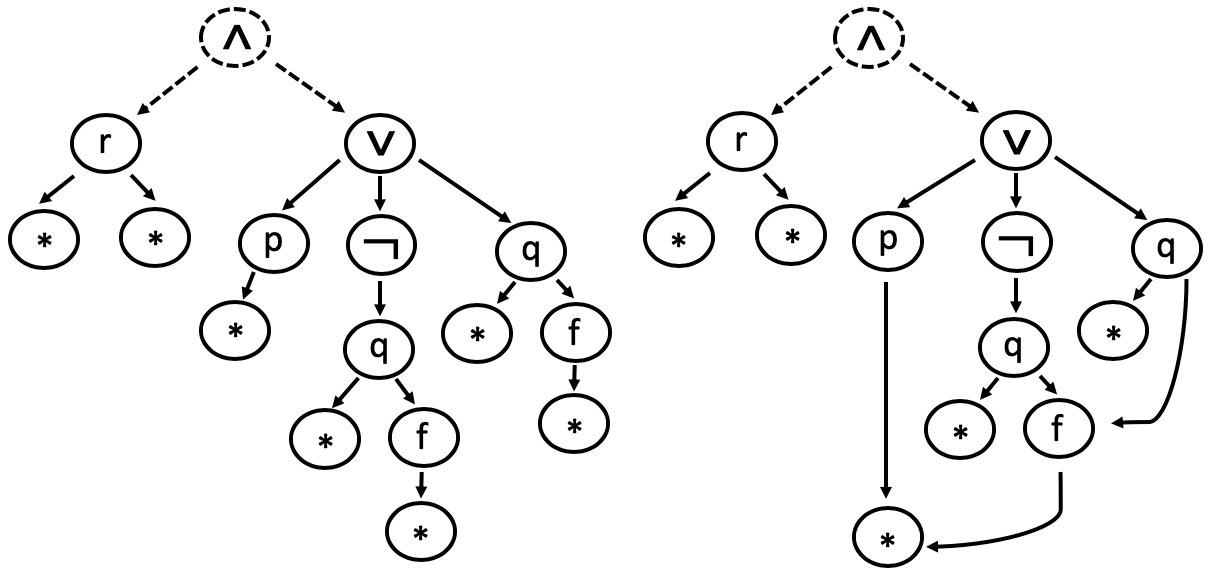}
\caption{A syntax-tree and a DAG representation of the formula $\forall A, B, C .\ r(A, B)\wedge\big(p(A) \vee \lnot q(B, f(A)) \vee q(C, f(A))\big)$. 
Only clause (sub)graphs are embedded in our system.
}
\label{fig:introformula}
\end{centering}
\end{figure}

As it is shown in Figure~\ref{fig:introformula} (left), the formula and its subexpressions are actually trees, and subsequent works \cite{JU-CICM17-enigma,Chvalovsky+-CADE19-enigmang,Crouse2019-trail} have taken this into account by developing patterns able to capture such structures. For instance, directed node paths oriented from the root, of length 3, and where variables are replaced by a placeholder symbol $*$, called term walks, 
are used as features in \cite{JU-CICM17-enigma,Chvalovsky+-CADE19-enigmang}.
More specifically, the embedding vector contains the counts of the feature occurrences so that, for our example, it would contain a 2 for the path ($q,f,*$) from $q$ to $A$. 
%
Also Tree LSTMs can be used for encoding the parse tree and they even allow for capturing it entirely, but they still only focus on the syntactic structure. 

For the actual logical interpretation (i.e., the semantics), formula characteristics like variable quantifications and shared subexpressions are quite important but are not captured by the syntax tree. Observe, for instance, that the syntax tree of the example contains several nodes for $A$, while, as mentioned above, all contexts $A$ occurs in should influence its interpretation. A simple iterative approach based on occurrences of $A$ also does not fully overcome the issue, since a variable $A$ may be interpreted differently in the context of different quantifiers/formulas.
For that reason, the most recent works focus on graph structures \cite{Crouse2019-premise-selection,olvsak2019property,paliwal2019graph}, and apply graph neural networks (GNNs) for encoding. Figure~\ref{fig:introformula} (right) depicts a graph representation with subexpression sharing.
However, while these sophisticated embeddings seem to be the most faithful to their input formulas, they also 
incur costs in terms of runtime, which might in turn result in an increased number of proof failures (given a time constraint).

It is not clear how these different encoding strategies compare against one another and which kind of strategy is best, or if there is such a strategy at all. 
When more advanced formula embeddings have been evaluated within the same ATP system, the evaluations did not consider similarly involved embeddings, but rather simple or very similar baselines which were often easier to implement.
In this paper, we conduct an experimental study on encodings of FOL formulas for automated theorem proving. The goal of this work is to provide an apples-to-apples comparison between encoding strategies by evaluating each of them within the same neural guided ATP, the TRAIL system \cite{Crouse2019-trail}.
Our results may help guide future research on neural guided ATPs.
In summary, our contributions are as follows:
\begin{itemize}
    \item We implemented 
    term walks \cite{JU-CICM17-enigma} and the pattern-based encoding proposed in \cite{Crouse2019-trail}; and several variants of graph neural networks 
    which, to the best of our knowledge, have not been considered in this context before,
    and integrated them into the TRAIL theorem prover.
    \item We evaluated the embeddings on the standard benchmarks Mizar \cite{grabowski2010mizar} and TPTP \cite{sutcliffe2009tptp}.
    \item We show that there is no single best-performing encoding, but that there are considerable differences in terms of runtime, completion rate, search effectiveness, and proof length, 
    some of which are rather unexpected.
\end{itemize}

The paper is organized as follows. Section~\ref{sec:related} gives an overview of related work, TRAIL is described in Section~\ref{sec:preliminaries}. The embeddings we compare are described in Sections~\ref{sec:approach1} and \ref{sec:approach2}, and the evaluation in Section~\ref{sec:evaluation}.
We assume the reader to be familiar with first-order logic and related concepts; see, e.g., \cite{taylor1999practical} for an introduction.

\section{Related Work}
\label{sec:related}

Most machine learning enhanced large-theory ATP systems extract symbol and structure-based features from their input formulas  (e.g., the depth or symbol count of a clause) 
in addition to hand-designed features known to be relevant to proof search (e.g., the age of a clause, referring to the time point when it was generated). While symbol-based features for a formula are generally just the multiset of symbols found in that formula, structure-based features vary in their design and implementation. Earlier large-theory ATP systems like Flyspeck \cite{kaliszyk2012learning} and MaLARea \cite{urban2008malarea} derived their features from the multisets of terms and sub-terms present in the formulas. These subsequently inspired the development of term walks (see 
Section~\ref{sec:approach1}) 
found in Mash \cite{kuhlwein2013mash} and Enigma \cite{JU-CICM17-enigma,Chvalovsky+-CADE19-enigmang}. In \cite{kaliszyk2015efficient}, pattern-based features were introduced from discrimination and substitution trees that could capture the notion of a term matching, unifying, or generalizing another term. In a similar vein, pattern-based features are also used in TRAIL, where the features extracted for a literal were argument-order preserving templates generated from the complete set of paths between the root and each leaf of the literal.

Recently, deep learning methods have demonstrated viability in the setting of real-time theorem proving, with the work of \cite{Loos+-LPAR2017-deep-nw-guided} being the first to show how a deep neural network could be incorporated into an ATP without incurring insurmountable computational overhead. Since then, there has been a flurry of activity \cite{Chvalovsky+-CADE19-enigmang,bansal2019holist,olvsak2019property,Crouse2019-trail} surrounding the applicability of neural networks in the ATP domain.
One of the main goals of these approaches is to also learn the formula embeddings in a fully automated way.

The latest developments for representing logical formulas as vectors have revolved around graph neural networks \cite{wang2017premise,paliwal2019graph,olvsak2019property}. These networks are appealing in the automated theorem proving domain because of the inherent graph structure of logical formulas and the potential for such neural representations to require less expert knowledge to achieve results than more traditional hand-crafted representations. Thus far, they have been applied in both offline tasks \cite{wang2017premise,Crouse2019-premise-selection}, and online theorem proving \cite{paliwal2019graph,olvsak2019property}. However, \cite{paliwal2019graph} focus on higher-order logic formulas where the corresponding graphs are very different from those for FOL. The latter were only considered very recently in \cite{olvsak2019property} for the first time.
\cite{wang2017premise} focus on a graph representation that only slightly extends parse trees by shared constant and variable names, while \cite{Crouse2019-premise-selection} extend them by special edges for quantification and subexpression sharing (see also Figure~\ref{fig:introformula} (right))\footnote{The work actually also introduces edges from quantifiers to variables which we do not show in Figure~\ref{fig:introformula} since, in our setting, we encode clauses and thus ignore these edges.}, and \cite{olvsak2019property} propose a special hypergraph representation. All the works use variants of message-massing neural networks \cite{gilmer2017neural} to learn and finally obtain a single numerical vector representation for each formula.

As of yet, little is known about the trade-offs between each of the aforementioned vectorization strategies as they would be used in a neural-guided theorem prover. This is in part due to some features not well lending themselves to the neural-guided theorem proving setting (e.g., features defined for full terms would be far too sparse, which is why recent approaches apply feature hashing \cite{Chvalovsky+-CADE19-enigmang}). 
The work of \cite{kaliszyk2015efficient} provides an extensive comparative analysis of various non-neural vectorization approaches, however, their evaluation focuses on sparse learners and it evaluates features in the setting of offline premise selection (measuring both theorem prover performance and binary classification accuracy) rather than as part of the internal guidance of an ATP system.
Very recently, \cite{gnn2019} presented a study comparing several graph neural networks for deciding FOL entailment and predicting proof length. While this study focuses on a similar goal as our paper -- and corroborates the need for this kind of work --, it compares the GNNs only to standard non-GNN architectures (e.g., LSTMs) but does not consider more involved pattern-based representations as used in state-of-the-art systems.



\section{The TRAIL Environment}
\label{sec:preliminaries}

TRAIL \cite{Crouse2019-trail} is an automated theorem proving environment in which the proof search is guided by reinforcement learning (RL). 
The proof search is a sequence of proof steps in which the set of input formulas (i.e., axioms and negated conjecture) is continuously extended by applying an action, which may lead to the derivation of new formulas, and stops if a proof (i.e., a contradiction) is found. An external FOL reasoner (whose proof guidance capabilities are suppressed) 
is used as the environment in which the learning agent operates. It tracks the state of the search and decides which actions are applicable in a given state.
The state encapsulates both the formulas that have been derived or used in the derivation so far and the actions that can be taken by the reasoner at the current step. At each step, this state is passed to the learning agent: an attention-based model~\cite{luong2015attention} that predicts a distribution over the actions and uses it to sample one action. This action is then given to the reasoner, which executes it and updates the proof state.

We focus on the representation of the formulas within TRAIL, i.e., in the proof state. All formulas are stored in conjunctive normal form as sets of \emph{clauses}, i.e., possibly negated atomic formulas called \emph{literals} which are connected via $\vee$. Our example contains the two clauses $\big(p(A) \vee \lnot q(B, f(A)) \vee q(C, f(A))\big)$ and $r(A, B)$, and \emph{positive} and \emph{negative} literals such as $p(A)$ and $\lnot q(B, f(A)) $.
The formula embedding approaches we study transform clauses into numerical vector representations.

\section{Pattern-Based Formula Embeddings}
\label{sec:approach1}
Pattern-based formula embeddings are usually based on the parse tree of the formula; see Figure~\ref{fig:introformula} (left) for our example. We evaluate \emph{term walks}, which have been used in \cite{JU-CICM17-enigma,LPAR-IWIL2018:ProofWatch_Meets_ENIGMA_First,Chvalovsky+-CADE19-enigmang}, and {chain patterns}, as representatives for patterns that capture entire literals (vs.\ parts of fixed depth or length).

\subsection{Term Walks}
As outlined in the introduction, literals in
the clauses are considered as trees where all variables and skolem terms 
are replaced by a special symbol, respectively.
Note that the latter helps reflecting structural similarities between literals that are indicative of unifiability. 
Additionally, a root node is added and labelled by either $\ominus$ or $\oplus$, depending on whether the 
literal 
appears negated or not. Every directed node path of length 3 in these trees (oriented from the root) represents a feature. 
%
%
For a negative literal such as $\lnot q(B, f(A))$, we would thus count term walks ($\ominus,q,*$), ($\ominus,q,f$), and ($q,f,*$).
The multiset of features for a clause consists of all features of its literals; and the final embedding vector for the clause has the same size as this set, every position is associated to one feature, and contains the multiplicities of the features at the corresponding positions.

\subsection{Chain Patterns}


The idea applied in TRAIL \cite{Crouse2019-trail} extends the simpler patterns of term walks in that
the clause embeddings should capture the literals more holistically (i.e., not only patterns of fixed depth), and the relationship between literals and their negations.
For example, in the term walks of our example, the first two  occurrences of $A$ are only captured by the term walk ($q,f,*$), so the connection to the contexts is largely lost although it might be rather important since there is a negation in the first one but not in the second. 

TRAIL patterns captures these features by deconstructing input clauses into sets of \emph{chain patterns}, where a pattern is a chain 
that begins from a predicate symbol and includes one argument (and its argument position) at each depth level until it reaches a constant 
or variable. The set of all chain patterns for a given literal is then simply the set of all linear paths between each predicate and the constants and variables they bottom out with. As with term walks, variables are replaced by a wild-card symbol $\ast$, 
and the latter is similarly used in all argument positions not in focus (i.e., not in the path under consideration). For our example, we obtain patterns such as $p(\ast),q(\ast,f(\ast)),q(\ast,\ast),$ and $r(\ast,\ast)$. Note that these patterns do not seem to differ much from term walks, but this changes when considering real-world problem clauses which are often much deeper than our toy example.
%

A $d$-dimensional clause embedding is obtained 
by hashing the linearization of each pattern $p$ using MD5 hashes 
to compute a hash value $v$, and setting the element at index $v \bmod d$ to the number of occurrences of the pattern $p$ in the clause under consideration. Further, the difference between patterns and their negations is explicitly encoded by doubling the representation size and hashing them separately, so that the first $d$ elements encode the patterns of positive literals and the second $d$ elements encode the negative ones. This hashing approach greatly condenses the representation size compared to one-hot encodings of all patterns.
%

\section{Graph-based Formula Embeddings} 
\label{sec:approach2}
\begin{figure}
\begin{centering}
\includegraphics[width=0.6\columnwidth]{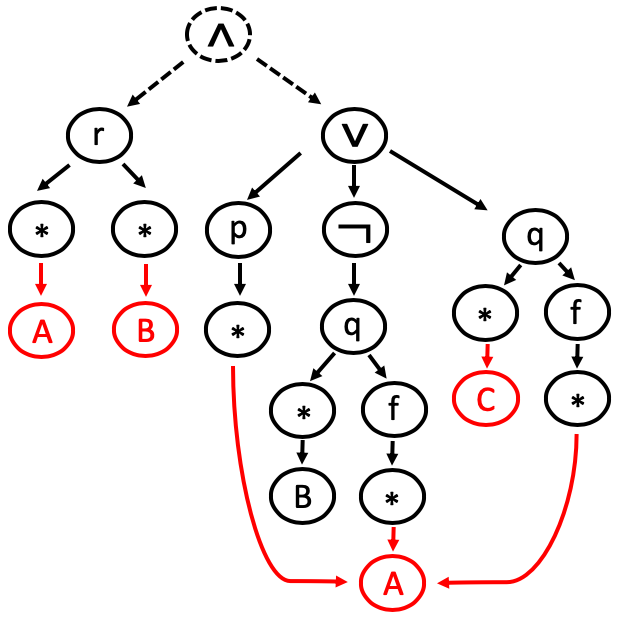}
\caption{The clause representation used in the GCNs.}
\label{fig:introformula2}
\end{centering}
\end{figure}

As outlined in the introduction, graph-based embeddings of logical formulas seem to better suit their actual semantics. However, such embeddings have been considered in the ATP context only very recently and focusing on the subtask of premise selection \cite{Crouse2019-premise-selection,olvsak2019property}. 
We consider the approach from \cite{Crouse2019-premise-selection}, which focuses on graphs as presented in Figure~\ref{fig:introformula} (right) (i.e., parse trees extended by subexpression sharing),
applies message-passing neural networks (MPNNs), and proposes a pooling technique to obtain the final graph encoding that has been specifically designed for formula graphs.
In addition, we evaluate variants of the simpler, but equally popular, graph convolutional neural networks (GCNs) in the ATP context. For the latter, we also use a relatively simple graph representation of formulas, depicted in Figure~\ref{fig:introformula2}, which only slightly extends the parse trees: in that variable and constant names are shared as suggested in \cite{wang2017premise} (vs. arbitrary subexpressions); note that we introduce \emph{name nodes} for this.

\subsection{
GCNs}

The originally proposed graph convolutional neural networks (GCNs) \cite{kipf2017gcn} compute node embeddings by iteratively aggregating the embeddings of neighbor nodes, and a graph embedding (i.e., a clause embedding) is obtained by a subsequent pooling operation, like max or min pooling.
The node embedding for a node $u$ is formalized as follows, assuming initial node embeddings $h_{u}^{0}$ are given:
\begin{align*}
h_{u}^{t+1} &= \rho\left(  
\sum_{v\in\mathcal{N}_{u}} 
{c_{u}}W_{r}^{t} h_{v}^{t} \right)
\label{eq:gcn}
\end{align*}
where $\mathcal{N}_{u}$ is the set of neighbors of node $u$, $c_{u}$ is a normalization constant,
$W^{t}$ is a learnable weight matrice, 
and $\rho$ is a non-linear activation function.

The initial node embedding can be obtained in various ways, and arbitrary initialization represents a common and easy solution. 
We additionally experimented with bag-of-character features (BoC), 
extracted without using any learning; and character features learned via a character convolutional neural network \cite{ZhangZL-NIPS15-charcnn}.
The idea behind these embeddings is to consider the names of the symbols in addition to the structural features of the formulas, which are encoded by the GCN.
Moreover, we do not want to rely upon a fixed token vocabulary, but to instead capture the overall shape of symbols which sometimes may encode important characteristics (e.g., in many datasets, variables or function symbols start with a fixed letter which is numbered). However, since the results for the character convolutional neural network turned out to be not competitive at all, we will omit them in our analysis later.

\subsection{Relational GCNs}
Relational GCNs (R-GCNs) \cite{schlichtkrull2018modeling} extend GCNs in that they distinguish different types of relations for computing node embeddings. Specifically, they learn different weight matrices for each edge type in the graph: 
\begin{align*}
h_{u}^{t+1} &= \rho\left(  
\sum_{r\in \mathcal{R}}
\sum_{v\in\mathcal{N}_{u,r}} 
{c_{u,r}}W_{r}^{t} h_{v}^{t} \right)
\end{align*}
Here, $\mathcal{R}$ is the set of edge types; 
$\mathcal{N}_{u,r}$ is the set of neighbors connected to node $u$ through the edge type $r$; $c_{u,r}$ is a normalization constant;
$W_{r}^{t}$ are the learnable weight matrices, one per $r\in \mathcal{R}$; 
and $\rho$ is a non-linear activation function. 

\subsection{MPNNs}
Message-passing graph neural networks (MPNNs) \cite{gilmer2017neural} 
extend GCNs in that the
aggregation of information from the local neighborhood includes edge embeddings: 
\begin{alignat*}{2}
&m_u^{t + 1} &&= \sum_{v \in \mathcal{N}_u} F^t_{M}([h_u^{t} ; h_v^{t} ; {e_{uv}}])  \\
&h_u^{t + 1} &&= h_u^{t} + F^t_{U}([h_u^{t} ; m_u^{t + 1} ])
\end{alignat*}
$F^t_{M}$ and $F^t_{U}$ are feed-forward neural networks 
and $[ \ \cdot \ ; \ \cdot \ ]$ denotes vector concatenation. 
The initial node embeddings $h_u^0$ and the edge embeddings ${e_{uv}}$ are assumed to be given for all nodes $u$ and $v$.
The vectors $m_u^{t}$ are \textit{messages} to be passed to $h_u$. 
As above, all the node embeddings from the last iteration are passed through a subsequent pooling layer, which computes the embedding for the whole graph. 



\subsection{DAG LSTM Pooling}

\begin{figure}
\begin{centering}
\includegraphics[width=0.6\columnwidth]{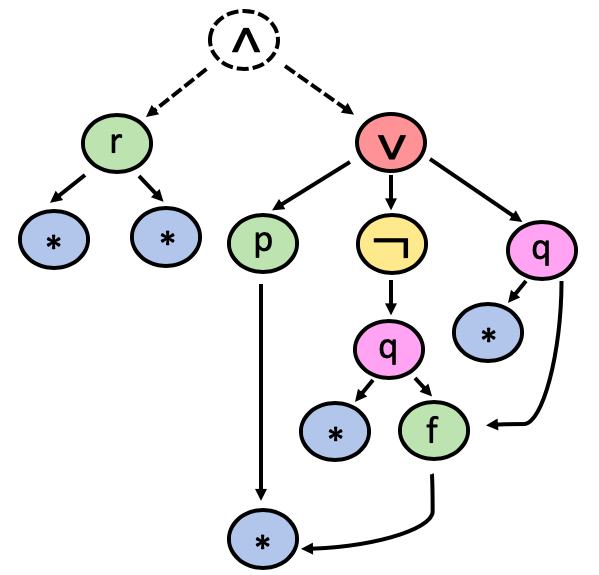}
\caption{Update staging in graph-based LSTM. Identically colored nodes are updated simultaneously, starting at leaves.
}
\label{fig:dag_upd}
\end{centering}
\end{figure}

In order to incorporate more of the information in the formula into its embedding,
\cite{Crouse2019-premise-selection} suggest to combine MPNNs with a pooling based on DAG LSTMs.
These LSTMs aggregate node and edge embeddings using techniques from LSTMs: 
input gates to decide which information is updated; 
tanh layers for creating the candidate updates;
memory cells for storing intermediate computations; 
forget gates modulating the flow of information from individual arguments into a node's computed state; and output gates similarly modulating the flow of information, but on a higher level.
Given initial node embeddings from the MPNN, and edge embeddings, new node embeddings are computed in topological order, starting from the leaves, as depicted in Figure~\ref{fig:dag_upd}. 
In this way, node embeddings are directly pooled and the embedding of the clause's root node represents the clause embedding.
For lack of space, we refer to the original paper for the exact computing equations.

\section{Evaluation}
\label{sec:evaluation}

In this section, we report the evaluation results of the embedding approaches described above. We also compare these approaches against Beagle~\cite{Beagle2015} (i.e., used inside TRAIL without suppressing its proving capabilities); a theorem proving system using manually designed heuristics which provides competitive performance on ATP datasets. 

\subsection{Network Configurations and Training}
We generally constructed embedding vectors of size 64 per clause with all approaches. 
The most important parameters for the individual approaches are described below.
%

For the GCNs, we used the parameters suggested in \cite{kipf2017gcn} (see Eq.~(2) and Sec.~3.1): (symmetric) normalized Laplacian as a normalization constant, ReLU activation, two convolutional layers in total, and Xavier initialization for the weights.
The output is passed through an additional linear layer and then pooled using summation, as suggested in \cite{xu2019gnnpower}.
We consider one GCN with arbitrary initial node embeddings (denoted GCN in the tables) and one based on initial bag-of-character embeddings (BoC-GCN).
The R-GCN implementation differs from the GCN in that we consider three edge types: edges to name nodes, edges from commutative operators to operands, all others. 

The MPNN configurations were taken from \cite{Crouse2019-premise-selection}. We considered one MPNN with max-pooling (MPNN) and one using the DAG LSTM (GLSTM-MPNN).

All our models were constructed in PyTorch\footnote{\url{https://pytorch.org/}} 
and trained with the Adam Optimizer \cite{kingma2014adam} with default settings. The loss function optimized for was binary cross-entropy. We trained each model for 5 epochs per iteration on all datasets. Validation performance was measured after each epoch and the final model used for the test data was then the version from the epoch with best performance.


\begin{table*}[t]
\centering
\resizebox{0.75\linewidth}{!}{%
\begin{tabular}{l|c|c|c|c|c|c}
\toprule
                           & \multicolumn{2}{c|}{Cumulative Compl.}&\multicolumn{2}{c|}{Best Iteration Compl.}&\multicolumn{2}{c}{Proof Length}  \\
                  &   Mizar    &   TPTP   &   Mizar    &   TPTP    &      Mizar  &    TPTP  \\
\midrule
Beagle (optimized v.)  &   63.3    &   42.0     & - & - &      1.00  &    1.00       \\
\hline
Term Walks                     &   \bf 52.7   &   24.3   & 43.2 &  20.1  &   \bf 1.01  &    0.10    \\
Chain Patterns                 &    51.5   &   \bf 32.0   &  \bf 50.3 &  \bf  28.9  &   0.48   &    \bf 0.12   \\
\hline
GCN                        &    39.6     &    21.9   & 40.8 & 18.3 &   0.50 &    0.10     \\
BoC-GCN                   &     43.8   &   24.3 & 40.2 &  20.1  &   1.25   &   \bf  0.14   \\
R-GCN                      &     42.0   &     20.7  & 42.0 & 17.16 &  \bf  2.13   &    0.11   \\
MPNN                       &    \bf 56.8   &  \bf  24.3  & \bf  51.5 &  \bf   23.1   &  1.89  &    0.11   \\
GLSTM-MPNN                      &    55.0   &    23.7  & 48.5 &  21.3 &  1.39  &    0.08   \\
\bottomrule
\end{tabular}
}
\caption{Performance of pattern and GNN-based encodings in terms of completion rate and proof length improvement relative to Beagle.}
\label{tab:comp_len}
\end{table*}



\begin{table*}[h]
\centering
\begin{tabular}{l|lll|lll|lll}
\toprule
         & \multicolumn{3}{|c}{Vectorization} & \multicolumn{3}{|c}{Action Selection} & \multicolumn{3}{|c}{Reasoning} \\
\midrule
         & Min      & Med     & Max       & Min   & Med  & Max    & Min     & Med    & Max     \\
\hline
Term Walks   & 0.00     & \bf 0.01       & 0.51      & 0.74  & \bf 0.43    & 9.98   & 0.19    & 3.22      & 72.82  \\
\rowcolor{lightgray}Chain Patterns & 0.10     & 0.04       & 2.53      & 0.03  & 0.70    & 15.86  & 0.18    & \bf 4.32      & 71.51  \\
GCN      &   0.02       &    0.68        &   78.40        & 0.03  & 0.80    & 78.52  & 0.08    & \bf 1.31      & 8.94   \\
BoC-GCN  &    0.03      &   1.43         &   69.65        & 0.04  & 1.59    & 70.03  & 0.14    & 2.87      & 27.27  \\
R-GCN    &  0.02        &   1.05         &   79.24        & 0.03  & 1.16    & 79.33  & 0.07    &  1.39      & 40.74  \\
\rowcolor{lightgray}MPNN     & 0.02     & 1.50       & 78.81     & 0.03  & 1.64    & 79.00  & 0.09    & 1.73      & 55.65  \\
GLSTM-MPNN    & 0.06     & \bf 2.65       & 81.39     & 0.07  & \bf 2.76    & 81.97  & 0.10    & 2.00      & 48.72  \\
\bottomrule
\end{tabular}
\caption{Average time spent per problem on Mizar for each phase. Minimal and maximal median numbers are in bold, best in completion rate are in gray.}
\label{tab:runtime}

\end{table*}

\subsection{Datasets and Experimental Setup}
We considered the standard Mizar\footnote{\url{https://github.com/JUrban/deepmath/}} \cite{grabowski2010mizar} and the Thousands of Problems for Theorem Provers (TPTP)\footnote{\url{http://tptp.cs.miami.edu/}} datasets. 
Mizar is a well-known and large mathematical library of formalized and mechanically-verified mathematical problems. TPTP is the definitive benchmarking library for theorem provers, designed to test ATP performance across a wide range of problem domains. 
From each dataset, 500 problems were drawn randomly. We used a 50/15/35 split for train/validation/test, and set a time limit of 100 seconds per problem solving attempt for each vectorization approach (thereafter the proof attempt was stopped and the problem declared unsolved). For training, we ran all models for 30 iterations over the training sets.

We consider the following metrics: 1)~\textit{Cumulative completion rate}~\cite{BaLoRSWi-CoRR19-learning}: this metric reports the proportion of problems solved across all testing iterations within the specified time limit.
2)~\emph{Best iteration completion performance}~\cite{kaliszyk2018reinforcement}: the ratio of problems solved at the best performing iteration. 
3)~\emph{Search efficiency}: the number of actions considered by TRAIL (executed and available actions). Among those, we also report the percentage of \emph{useless steps} which is the percentage of actions executed that were not used directly in the derivation of the final contradiction.
4)~\emph{Relative proof length}: the average proof length across all problems which is measured as the number of proof steps found by TRAIL divided by the length of proof found by Beagle (a value greater than one indicates a shorter proof compared to Beagle's). 5)~\emph{Runtime}: for different phases of problem solving. 

\subsection{Results and Discussion}

\paragraph{Completion Rate.} We show in Table~\ref{tab:comp_len} the cumulative and the best iteration's completion rate of each embedding approach on both Mizar and TPTP datasets. Table~\ref{tab:comp_len} 
shows that there is a  gap between the completion rate of Beagle and the approaches we evaluate. Note that this gap is larger than in other ATP evaluations, however, many of the latter include symbol or structure-based features (while we solely consider the encodings learned) or compare to ATPs instead of to manually-designed systems. Cumulatively, MPNN and GLSTM-based embeddings solved more problems than all other approaches on Mizar dataset while chain patterns solved more problems on TPTP dataset. The performance of the best iteration still shows that MPNN and chain-based embeddings perform the best compared to the other techniques. 
%
%
Specifically, the chain-based patterns, which capture the formulas more holistically than term walks turn out to largely outperform the latter. 
The GCN variants, which are combined with a simpler graph representation of formulas, generally perform worse that the pattern-based encodings.
On the other hand, the message-passing neural networks outperform the term walks.
In our experiments, the GLSTM pooling does not provide benefits over a standard max-pooling with the message-passing neural network.
MPNN slightly outperforms the chain  patterns on Mizar, but it is the opposite on TPTP.

\paragraph{Search Efficiency.}

\begin{figure}[t]
\begin{centering}
\includegraphics[width=1\columnwidth]{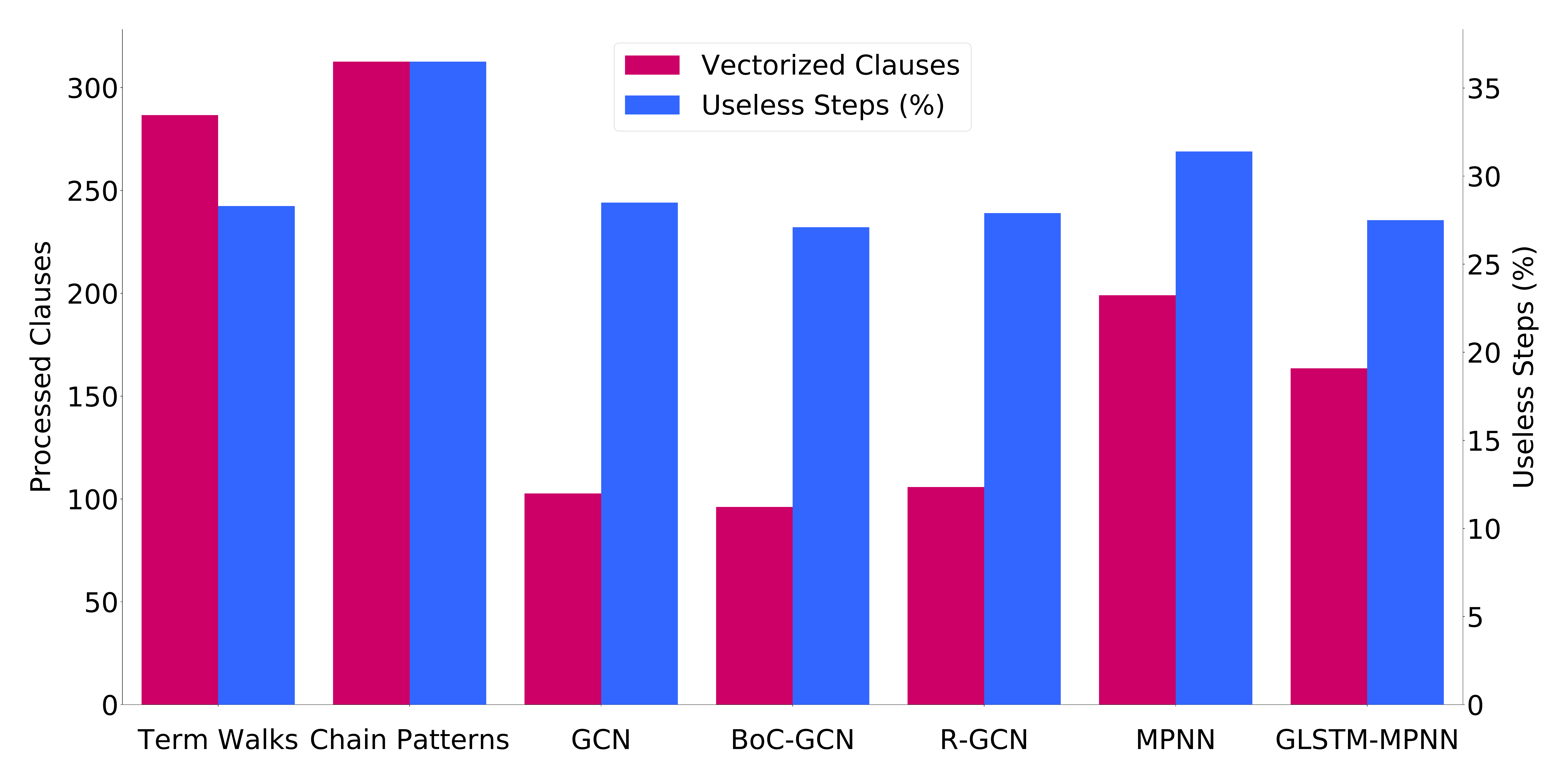}
\caption{Average number of vectorized clauses per problem. }
\label{fig:clauses}
\end{centering}
\end{figure}

In order to get a sense of how efficiently proof search was performed using each vectorization strategy, we examine the average number of actions considered per problem. This number counts both actions taken by TRAIL as well as those actions that were available but ultimately remained unexplored. Both taken and untaken actions are included because all actions, whether taken or not, are vectorized (since the learning agent uses a neural network to predict if they are to be taken or not), and thus incur runtime overhead in TRAIL. Figure \ref{fig:clauses} shows this number for each of the vectorization approaches. Notably, in the figure we see that the simpler pattern-based embeddings resulted in substantially more actions considered than the MPNN and GLSTM-MPNN approaches. This difference is likely the source of the effectiveness for both the MPNN and GLSTM-MPNN which seem to have a more concise search space and hence better chances in finding a proof.  

We further support our claim regarding conciseness by measuring the percentage of useless steps. This is calculated as the number of executed actions that were \emph{not} returned in the final proof divided by the total number of actions taken by TRAIL. This metric is somewhat biased, as an approach that solves fewer problems where the problems are easier (i.e., shorter proof required), will achieve better scores. However, it is interesting as a comparison point for those approaches that performed similarly well in terms of completion rate. Noteworthy to mention is that the chain patterns approach led to the highest percentage of useless steps, which would seem to indicate a more breadth-first approach to problem solving. Conversely, that the MPNN had both shorter proofs and a lesser fraction of useless steps would seem to indicate that the MPNN is exploring in a more targeted depth-first fashion.

\paragraph{Proof Length.}
We also report in Table~\ref{tab:comp_len} the average proof length of each approach.
We measure proof length as it is considered to be an indicator of proof simplicity \cite{veroff2001finding,wos2001hilbert,kinyon2019proof}, with shorter proofs being considered simpler than longer ones.
While the length of proofs found should not be considered as a criterion as important as completion rate, it may influence the choice of embedding in specific application scenarios, e.g., when proof outputs must be inspected or utilized by human end-users.
On Mizar, most graph-based encodings lead to much shorter proofs than their pattern-based counterparts (for the problems they could solve) with GCNs finding the shortest proofs (confirmed by further experiments) followed by MPNN and others; larger values indicate shorter proofs. 


\paragraph{Runtime.}
Table~\ref{tab:runtime} shows the average runtimes for the three main phases of proving a problem; 1) vectorization: time spent in encoding the logical formulas, 2) action selection: time spent in evaluating the RL policy network for selecting the next action, and 3) reasoning: time spent in executing the selected action and producing the next system state. As expected, pattern-based  encoders require less time on average than graph-based approaches for encoding the logical formulas and evaluating the policy network, which allows them to spend more time on reasoning. 



\paragraph{Discussion.}

The primary goal of research into neural-guided ATPs is to obviate the need of hand-crafted features and patterns, and recent advances seem to agree that GNNs are the way forward.
Our experiments have confirmed the direction taken by the first existing ATP works on GNNs, which focus on MPNNs instead of GCNs. Our particular MPNNs are costly in terms of runtime, however the implementations are not optimized with the latest libraries for GNN development, thus we would expect to see improvement in this regard.

To us, it came at surprise that the hand-crafted but rather simple patterns are still competitive given the broad effectiveness of graph neural networks throughout other domains.
This is likely due to the former being very performant in terms of encoding time as well as partially capturing properties like argument order and unifiability.
Automated theorem proving is a particularly hard application domain for GNNs and our analysis has shown that straightforward graph representations of formulas are not sufficient to achieve good performance in general. Very recently, there has been an effort to encode more of the logic into the graphs and correspondingly adapt the GNN \cite{olvsak2019property}. Although the latter evaluation only compares to one other ATP system, we believe that this kind of encoding is the direction to take and needs further investigations. 

\section{Conclusions}
\label{sec:conclusions}

Until now, there had been little work investigating the trade-offs between the different embedding strategies for FOL 
formulas in automated theorem proving. In this paper, we presented an experimental study comparing the performance of various such strategies in the context of the TRAIL system, thus allowing for a fair, direct comparison to be made between embedding types.
We implemented two pattern-based approaches and several variants of graph convolutional and message-passing neural networks, and thus considered a representative set of several popular and recent standard graph embedding methods varying in complexity. As in prior work, we evaluated 
them in terms of completion rate, but we also went further to give a detailed analysis with regards to search efficiency, proof length, and runtime as well. By highlighting the strengths and deficiencies of a wider range of vectorization approaches, our findings should be more broadly useful to those seeking to improve their own neural-guided ATP systems, regardless of the vectorization strategy they employ.


\bibliographystyle{kr}
\bibliography{refs}

\end{document}